\pdfoutput=1

\documentclass[11pt]{article}

\usepackage[final]{acl}  

\usepackage{times}
\usepackage{latexsym}
\usepackage{hyperref}
\usepackage[T1]{fontenc}

\usepackage[utf8]{inputenc}

\usepackage{microtype}
\usepackage{booktabs}
\usepackage{tabulary}
\usepackage{xcolor}
\usepackage{colortbl}
\usepackage{ulem}
\usepackage{array}

\usepackage{amsmath}        
\usepackage{amssymb}        
\usepackage{graphicx}       
\usepackage{algorithm}      
\usepackage{algorithmicx}     
\usepackage{algpseudocode}  
\usepackage{hyperref}       
\usepackage{cleveref}       
\usepackage{amsthm}
\usepackage{siunitx}
\usepackage{multirow}
\usepackage{makecell}

\definecolor{lightblue}{RGB}{235,245,255}
\definecolor{highlight}{RGB}{230,230,230}

\usepackage{inconsolata}

\usepackage{graphicx}

\algnewcommand{\algorithmicinput}{\textbf{Input:}}
\algnewcommand{\algorithmicoutput}{\textbf{Output:}}
\algnewcommand{\INPUT}{\item[\algorithmicinput]}
\algnewcommand{\OUTPUT}{\item[\algorithmicoutput]}

\title{Low-Confidence Gold: Refining Low-Confidence Samples for Efficient Instruction Tuning}

\author{
 Hongyi Cai\textsuperscript{1,*} \quad
 Jie Li\textsuperscript{2,*} \quad
 Mohammad Mahdinur Rahman\textsuperscript{1} \quad
 Wenzhen Dong\textsuperscript{3}
 \\
 \textsuperscript{1}Universiti Malaya \quad
 \textsuperscript{2}University of Science and Technology Beijing \quad
 \textsuperscript{3}The Chinese University of Hong Kong
 \\
 {\small \textsuperscript{*}Corresponding: \href{mailto:xcloudfance@gmail.com}{xcloudfance@gmail.com}}
}

\begin{document}
\maketitle
\begin{abstract}
The effectiveness of instruction fine-tuning for Large Language Models is fundamentally constrained by the quality and efficiency of training datasets. This work introduces Low-Confidence Gold (LCG), a novel filtering framework that employs centroid-based clustering and confidence-guided selection for identifying valuable instruction pairs. Through a semi-supervised approach using a lightweight classifier trained on representative samples, LCG curates high-quality subsets while preserving data diversity. Experimental evaluation demonstrates that models fine-tuned on LCG-filtered subsets of 6K samples achieve superior performance compared to existing methods, with substantial improvements on MT-bench and consistent gains across comprehensive evaluation metrics. The framework's efficacy while maintaining model performance establishes a promising result for efficient instruction tuning. All open-source assets are publicly available at \url{https://github.com/Lizruletheworld/Low-Confidence_Gold}.
\end{abstract}
\section{Introduction}

Large Language Models (LLMs) have been trained to follow instructions by specific supervised response data after pre-training stage. Many instruction finetuning (IFT) \cite{alpaca} datasets emerge to realize various downstream tasks, for example: mathematic calculation, sentence analysis, haiku writing and etc, aiming to strengthen the ability of LLMs in instruction following. To save vast human costs for data annotation, most of studies introduce other teacher LLMs (e.g. \texttt{text-davinci-003} \cite{gpt}) to align the best instructions with corresponding responses. 

However, IFT datasets (e.g. \texttt{Alpaca\_52k} \cite{alpaca}, \texttt{magpie} \cite{xu2024magpie}) suffer from misleading content and poor quality, resulting in the bottleneck of post-training performance, even though teacher models replenish the missing parts of context and instruction pairs. This highlights the need for effective data filtering methods that identify high-quality instruction subsets while reducing fine-tuning time and computational costs.

\begin{figure}[t]
    \centering
    \includegraphics[width=0.9\linewidth]{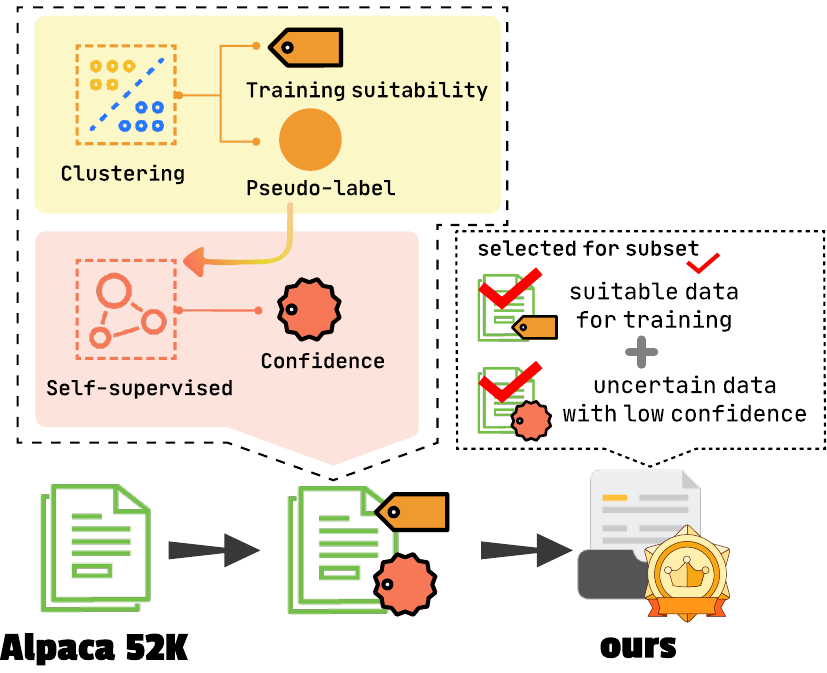}
    \caption{We target to select complex and quality samples confidence ranking for benefiting LLM training.}

    \label{fig:teaser}
\end{figure}

\texttt{Alpagasus} \cite{chen2023alpagasus} proposed a model-based approach that introduces proprietary LLMs to score data quality in multiple facets, replacing human annotation by taking advantage of the automated pipeline. However, this leads to datasets that are likely biased by the preference for redundant and limited responses \cite{panickssery2024llm, mergeit}, which potentially deteriorates the diversity of the original data. \citealp{ge2024clustering} emphasizes the necessity of diversity and therefore proposed clustering and ranking to select subsets of data. Further, Superfiltering \cite{li2024superfiltering} gains more insights in small open-source LLM that scores the instruction following ability of \texttt{Alpaca\_52k}. Although the instruction score provides an efficient and simple criterion for data selection, it does not consistently correlate with both the quality and diversity of data. Consequently, improvements in performance may not always be guaranteed.

To address these challenges, we propose a novel data filtering framework, \textbf{Low-Confidence Gold (LCG)} for efficient instruction tuning that significantly reduces computational costs while maintaining model performance. Our approach, shown in \ref{fig:teaser}, innovatively seeks to identify high-value instruction data through classification tasks. Specifically, we develop a lightweight classification model trained on centroid subsets that effectively categorizes instruction-response pairs, and leverage low-confidence predictions to curate challenging examples most beneficial for instruction tuning. Another perspective is that, since the common instruction tuning data are lack of annotations and labels, we adopt the manner of semi-supervised learning, to construct pseudo-labels as our training groundtruth, as well as getting inspired quality data from affordable yet effective models.

Through extensive experiments on the \texttt{Alpaca\_52K dataset}, we demonstrate that our filtered subsets achieve comparable or better performance when fine-tuning various open-source language models, while requiring only a fraction of the original data. Our main contributions are threefold: 
\begin{enumerate}
    \item A novel and efficient data filtering paradigm for instruction tuning that combines nearest neighbor classification with confidence-based selection.
    \item We train a small classifier model that enables selection for the whole set of instruction finetuning data.
    \item Experiments and evaluations are conducted that demonstrate the outstanding effectiveness of our filtered datasets working on multiple open-source LLMs. We reach \textbf{states-of-the-arts performance} in MT-Bench and HuggingFace OpenLLM Leaderboard benchmarks.
\end{enumerate}

\section{Preliminaries}

\subsection{K-means Clustering}

Given the Alpaca\_52k dataset $\mathcal{D} = \{(x_i, y_i)\}_{i=1}^{N}$ where $N=52,000$, we first cluster instructions into $K$ semantic groups using K-means. Let $\phi(x_i) \in \mathbb{R}^d$ denote the embedding vector of instruction $x_i$. The clustering objective minimizes:

\begin{equation}
\min_{\{C_k\}_{k=1}^K} \sum_{k=1}^K \sum_{x_i \in C_k} \|\phi(x_i) - \mu_k\|^2
\end{equation}

where $\mu_k = \frac{1}{|C_k|}\sum_{x_i \in C_k} \phi(x_i)$ is the centroid of cluster $C_k$. This partitions $\mathcal{D}$ into $K$ disjoint subsets $\{C_1, ..., C_K\}$ based on instruction similarity.

\subsection{Problem Setting}
Our filtering framework, LCG aims to select a subset $\mathcal{D}_{\text{filtered}} \subseteq \mathcal{D}$ that satisfies:

\begin{equation}
\mathcal{D}_{\text{filtered}} = \bigcup_{k=1}^K \left\{ (x_j, y_j) \in C_k \mid \mathcal{F}(x_j, y_j) < \tau_k \right\}
\end{equation}

where $\mathcal{F}: \mathcal{D} \rightarrow [0,1]$ is a discriminative confidence scorer and $\tau_k$ is an adaptive threshold for cluster $C_k$. The scorer $\mathcal{F}$ evaluates how "hard" a sample is to be trivially categorized, with higher values indicating the simplicity of data which is easily determined and differentiated. The training efficiency therefore increases since only a small subset of instructions are curated.

\begin{figure*}[t]
  \includegraphics[width=1.0\linewidth]{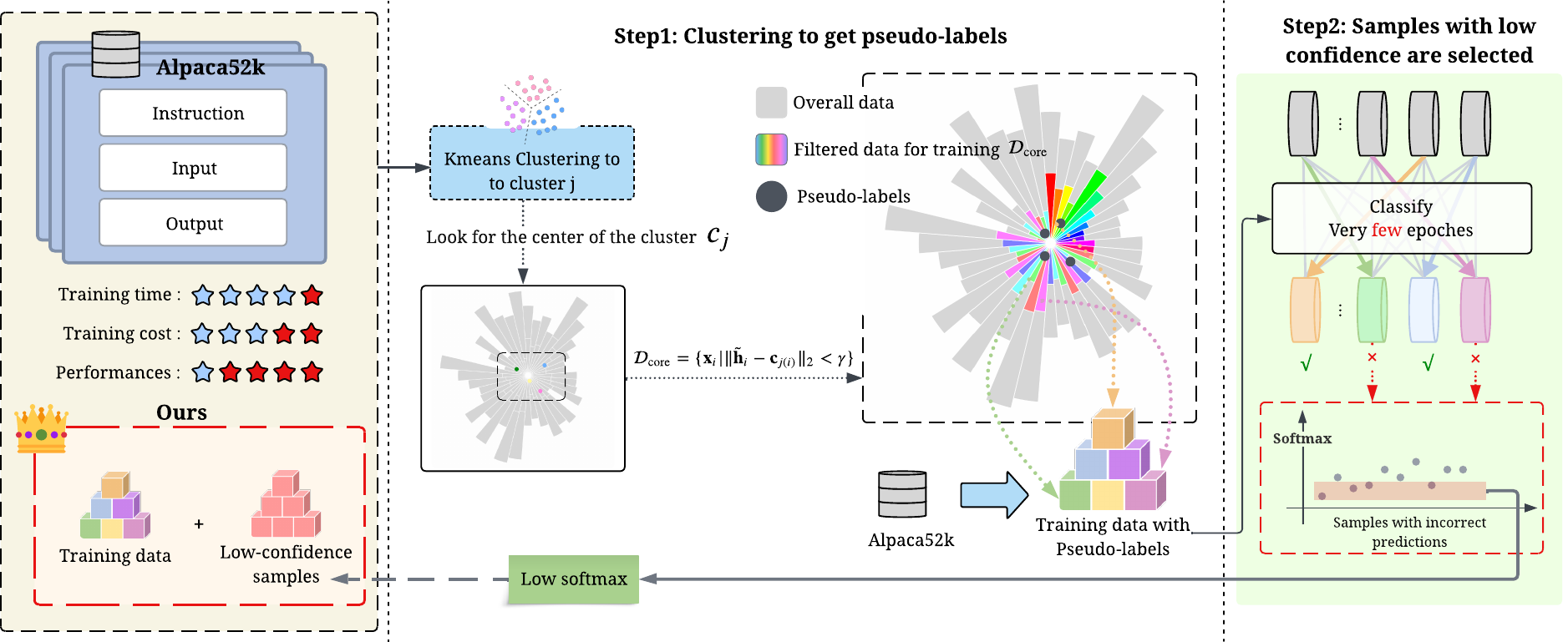}
  \caption{The overall pipeline of Low-Confidence Gold. We split our pipeline into two main steps: 1) Clustering to get pseudo-labels and centroid data to collect the initial diversity of data. 2) We feed annotated data into a tiny yet effective classifier to rank the confidences for the rest of the distant data to implement subset selection. }
\end{figure*}

\section{Methodology}
\label{sec:method}

\subsection{Motivation}
\textbf{Instruction filtering demands a dual-focus mechanism that intrinsically balances data quality and diversity.} Traditional supervised methods face inherent scalability limitations as manual annotation becomes prohibitively expensive for large-scale instruction datasets \cite{liu-etal-2022-makes, longpre2023flan, deita}. Meanwhile, it is difficult to identify suitable and challenging data for LLMs training without introducing proprietary LLMs or labors. Our semi-supervised framework addresses these limitations through pseudo-label refinement and early-stopped confidence detection, creating dynamic selection boundaries aligned with language model learning dynamics.

\textbf{Cluster-centric pseudo-labeling addresses data distribution challenges in instruction tuning.} Traditional sampling methods often struggle to balance between common and rare instruction patterns, leading to either over-representation of frequent cases or loss of valuable rare examples. We create semantic clustering anchors that naturally preserve the diversity of instruction patterns. By sampling 3\% of data points nearest to cluster centroids, we ensure each semantic category contributes meaningful examples while maintaining the inherent data distribution characteristics.

\textbf{Early-stopped classifier training induces uncertainty to identify high-quality samples.} Limiting the classifier to 3 epochs creates deliberate underfitting - the model develops basic pattern recognition without over-specializing to pseudo-labels. When applied to non-centroid samples, this partially-trained classifier's low-confidence predictions signal instructions containing non-trivial semantic constructs. These samples challenge the classifier's emerging decision boundaries precisely because they contain valuable complexity that language models should master, not avoid.

\subsection{Centroid Coreset Selection for Pseudo-labels}
In the initial step of our approach, we select a coreset from the whole corpus to identify pseudo-labels by the K-means algorithm, which effectively determine each semantic clusters. Given a dataset of instruction pairs $\mathcal{D} = \{(\mathbf{x}_i, \mathbf{y}_i)\}_{i=1}^N$, we first encode each instruction $\mathbf{x}_i$ into a dense vector representation using MiniLM \cite{wang2022minilmv2}:

\begin{equation}
\mathbf{h}_i = \text{AvgPool}(\text{MiniLM}(\mathbf{x}_i)) \in \mathbb{R}^{384}
\end{equation}

This geometric progression ensures proportional coverage of both frequent and rare instruction patterns. Cluster centroids $\{\mathbf{c}_j\}_{j=1}^k$ are computed via:

\begin{equation}
\mathbf{c}_j = \frac{1}{|\mathcal{C}_j|} \sum_{\mathbf{x}_i \in \mathcal{C}_j} \tilde{\mathbf{h}}_i
\end{equation}
where $\mathcal{C}_j$ denotes the set of samples assigned to cluster $j$. Centroid-proximal samples are selected as high-confidence candidates:

\begin{equation}
\mathcal{D}_{\text{core}} = \{\mathbf{x}_i | \|\tilde{\mathbf{h}}_i - \mathbf{c}_{j(i)}\|_2 < \gamma\}
\end{equation}
where $\gamma$ is the 90th percentile distance within each cluster.

\subsection{Low-Confidence Gold: Calibrating with Low-confidence samples to select data}

After determining pseudo-labels based on clusters, those annotations can be served for classification training. Specifically, we train a multi-class classifier on the core samples $\mathcal{D}_{\text{core}}$. The model architecture consists of:

\begin{equation}
f_\theta(\mathbf{x}) = \text{Softmax}(\mathbf{W}_2 \cdot \text{GELU}(\mathbf{W}_1 \mathbf{h}_i + \mathbf{b}_1) + \mathbf{b}_2)
\end{equation}

where $\mathbf{W}_1 \in \mathbb{R}^{384 \times 768}$, $\mathbf{W}_2 \in \mathbb{R}^{768}$ are learnable parameters, and GELU denotes the Gaussian Error Linear Unit activation. The model optimizes cross-entropy loss:

\begin{equation}
\begin{aligned}
\mathcal{L}(\theta) = -\frac{1}{|\mathcal{D}_{\text{core}}|} 
\sum_{(\mathbf{x}_i,y_i)} & \sum_{j=1}^k \mathbb{I}(y_i=j) \\
& \cdot \log p_\theta(y=j|\mathbf{x}_i)
\end{aligned}
\end{equation}

Training terminates at epoch $T=3$ since we aim to keep the model in an early-stopped stage so that they would not overfit to the centroid subset data. After training, we rank the confidence distribution calculated from \textit{softmax} function and select the top $K$ most uncertain data in each cluster.

\begin{table*}[t!]
    \centering
    \scalebox{0.92}{
    \definecolor{lightblue}{RGB}{235,245,255}
    \definecolor{highlight}{RGB}{230,230,230}
    \begin{tabulary}{\textwidth}{L|C|CCCCC}
        \toprule
        \rowcolor{lightblue} 
        \textbf{Model} & \textbf{MT-bench} & \multicolumn{5}{c}{\textbf{Huggingface Open LLM Leaderboard Scores (\%)}} \\
        \cmidrule{3-7}
        \rowcolor{lightblue} 
        & \textbf{Score} & \textbf{Hellaswag} & \textbf{MMLU} & \textbf{GSM8k} & \textbf{ARC} & \textbf{Avg} \\
        \midrule
        
        \multicolumn{7}{l}{\textit{First Group - Base Model}} \\
        Mistral-7b-v0.3 & 3.639 & 60.94 & 58.96 & 36.62 & 48.81 & 51.33 \\
        \midrule
        
        \multicolumn{7}{l}{\textit{First Group - Methods}} \\
        Alpaca-52k & 4.018 & 61.18 & 57.73 & 31.61 & 53.07 & 50.90 \\
        SuperFiltering-10\% & 3.963 & 60.98 & 59.34 & 35.71 & 49.83 & 51.47 \\
        Random-6k & 4.314 & 60.83 & 58.75 & 35.03 & 53.07 & 51.92 \\
        Perplexity-6k & 4.352 & 61.64 & 58.48 & 37.00 & 51.88 & 52.25 \\
        Kmeans-6k & 4.283 & 60.86 & 58.45 & 35.10 & 52.05 & 51.62 \\
        LIMA-6k & 4.440 & 60.58 & 59.34 & 37.31 & 51.11 & 52.09 \\
        \rowcolor{highlight}
        \textbf{LCG-MultinomialNB-6k (Ours)} & \textbf{5.086} & \textbf{62.00} & \textbf{59.51} & \textbf{40.51} & \textbf{52.90} & \textbf{53.73} \\
        \rowcolor{highlight}
        \textbf{LCG-DistilBERT-6k (Ours)} & \uwave{4.894} & \uwave{61.99} & \textbf{59.51} & \uwave{40.33} & \uwave{52.22} & \uwave{53.51} \\
        \rowcolor{highlight}
        \textbf{LCG-DistilBERT-1k (Ours)} & 4.869 & 61.94 & \uwave{59.24} & 38.29 & 51.62 & 52.77 \\
        \midrule
        
        \multicolumn{7}{l}{\textit{Second Group - Base Model}} \\
        LLaMa3-8b & 3.418 & 60.17 & 62.13 & 50.42 & 50.26 & 49.98 \\
        \midrule
        
        \multicolumn{7}{l}{\textit{Second Group - Methods}} \\
        Alpaca-52k & 3.718 & 60.57 & 61.36 & 46.10 & 52.41 & 55.74 \\
        SuperFiltering-10\% & 3.968 & 60.38 & 61.95 & \uwave{50.34} & 51.54 & 55.36 \\
        Random-6k & 3.912 & 60.83 & 58.75 & 35.03 & 53.07 & 51.92 \\
        Perplexity-6k & 4.120 & \uwave{61.14} & 61.09 & 50.87 & \uwave{53.50} & 56.65 \\
        Kmeans-6k & 3.731 & 60.86 & 58.45 & 35.10 & 53.07 & 51.87 \\
        LIMA-6k & 4.450 & 60.58 & \uwave{62.13} & 50.34 & 51.11 & 55.82 \\
        \rowcolor{highlight}
        \textbf{LCG-MultinomialNB-6k (Ours)} & \uwave{4.815} & \uwave{61.61} & 62.23 & \uwave{53.75} & \uwave{54.95} & \uwave{58.14} \\
        \rowcolor{highlight}
        \textbf{LCG-DistilBERT-6k (Ours)} & \textbf{4.963} & \textbf{61.43} & \textbf{62.67} & \textbf{54.28} & \textbf{54.78} & \textbf{58.29} \\
        \rowcolor{highlight}
        \textbf{LCG-DistilBERT-1k (Ours)} & 4.776 & 60.95 & 62.26 & 52.92 & 52.82 & 57.23 \\
        \bottomrule
    \end{tabulary}
    }
    \caption{Performance comparison on standard benchmarks. Results in \textbf{bold} indicate best performance within each group, while \uwave{underlined} values represent second-best performance within each group. The table is divided into two groups, each with its base model and various fine-tuning methods. We add the complete results for LCG-MultinomialNB-6k on LLaMa3-8b.}
    \label{tab:model_comparison}
\end{table*}

\section{Experiments}
\label{sec:experiments}

\subsection{Experimental Setup}
We utilize LCG to filter the \texttt{Alpaca\_52k} dataset. Two classifiers are used for the selection process: Multinomial Naive Bayes (MNB) \cite{scikit-learn} and DistilBERT \cite{sanh2019distilbert}. The classifiers are trained for 3 epochs with a learning rate of 1e-5. We then select curated datasets by a confidence threshold of $<0.7$. The resulting subsets are used to fine-tune two open-source LLMs: \texttt{Mistral-7b-v0.3} \cite{jiang2023mistral} and \texttt{LLaMa3-8b} \cite{dubey2024llama}, using LoRA \cite{hu2021lora} with a learning rate of 2e-5 for 3 epochs.

\begin{table*}[t!]
\centering
\scalebox{0.85}{
\begin{tabular}{lccccc}
\toprule
\rowcolor{lightblue}
\textbf{Models (Mistral-7B)} & \textbf{ARC} & \textbf{GSM8k} & \textbf{HellaSwag} & \textbf{MMLU} & \textbf{Avg} \\
\midrule
WizardLM-50k (Full) & 51.54 & 38.59 & 62.14 & 59.36 & 52.91 \\
WizardLM-Longest & 51.08 & 38.99 & 62.10 & 58.99 & 52.79 \\
WizardLM-Perplexity & 51.19 & 39.77 & 62.21 & 59.01 & 53.05 \\
WizardLM-K-means & 51.96 & 39.95 & 62.11 & 59.38 & 53.35 \\
WizardLM-SuperFiltering & 51.54 & 39.65 & 62.21 & 59.22 & 53.16 \\
\midrule
\rowcolor{highlight}
\textbf{LCG-DistilBERT-1k (Ours)} & \textbf{52.31} & \textbf{40.00} & \textbf{62.39} & \textbf{60.01} & \textbf{53.68} \\
\rowcolor{highlight}
\textbf{LCG-MNB-6k (Ours)} & 51.67 & 38.97 & 62.28 & 60.75 & 53.42 \\
\bottomrule
\end{tabular}
}
\caption{Cross-dataset validation on WizardLM with Mistral-7B. Our LCG method demonstrates superior performance against baselines.}
\label{tab:wizardlm_results}
\end{table*}

\subsection{Main Results}
As presented in Table \ref{tab:model_comparison}, our proposed LCG method consistently outperforms existing instruction data filtering approaches. When applied to Mistral-7b, LCG with MultinomialNB achieves the highest MT-bench score of 5.086, surpassing the previous best (LIMA-6k \cite{LIMA}) by 14.5\%. Similarly, LCG with DistilBERT demonstrates superior performance on LLaMA3-8b, improving the MT-bench score by 11.5\% over LIMA-6k. Notably, our method maintains strong performance even with only 1k examples, highlighting its effectiveness. The consistent improvements across diverse metrics (Hellaswag \cite{hellaswag}, MMLU \cite{mmlu}, GSM8k \cite{gsm8k}, and ARC \cite{arc}) further validate the robustness of our approach.

Our LCG-tuned models show particularly strong performance on MT-Bench. We attribute this success to the fact that low-confidence samples often represent nuanced dialogue scenarios, edge cases, and diverse response patterns. Training on these samples equips the model with superior reasoning capabilities, essential for the sophisticated multi-turn conversations in MT-Bench.

\section{Analysis and Discussion}

\subsection{Cross-Dataset Generalization}
To validate the robustness of our approach, we applied LCG to the \texttt{WizardLM} dataset, which is larger and more complex than Alpaca. As shown in Table \ref{tab:wizardlm_results}, our method continues to outperform other selection strategies, confirming that LCG is not overfitted to a single dataset but offers a generally applicable mechanism for curating high-quality instruction data.

\subsection{Ablation Studies}

We conducted ablation studies on key hyperparameters to validate our methodological choices.

\paragraph{Impact of Confidence Threshold $\tau$} The confidence threshold $\tau$ directly controls the size of the filtered subset. As shown in our main results (Table~\ref{tab:model_comparison}), relaxing $\tau$ to expand the dataset from 1K to 6K samples consistently improves performance on reasoning tasks like GSM8k. However, further experiments with a looser threshold (yielding 9K samples) resulted in significant performance degradation, indicating that our moderate threshold effectively balances quality and quantity by filtering out noise.

\subsection{Synergy with RLHF}
To explore modularity, we designed a two-stage pipeline combining LCG with Reinforcement Learning from Human Feedback (RLHF). First, LCG selects 10K candidates. Then, an RLHF preference model refines this set to the top 6K. As shown in Table \ref{tab:rlhf_results}, this combined approach significantly improves performance, especially on reasoning tasks, demonstrating that LCG serves as an effective initial filter for more advanced refinement techniques.

\begin{table}[h!]
\centering
\scalebox{0.75}{
\begin{tabular}{lcccc}
\toprule
\rowcolor{lightblue}
\textbf{Method} & \textbf{GSM8k} & \textbf{HellaSwag} & \textbf{MMLU} & \textbf{Avg} \\
\midrule
RLHF-Only-6k & 39.85 & 61.45 & 60.12 & 53.84 \\
\rowcolor{highlight}
\textbf{LCG+RLHF-6k} & \textbf{42.15} & \textbf{62.35} & \textbf{60.45} & \textbf{54.66} \\
\bottomrule
\end{tabular}
}
\caption{Performance of combining LCG with RLHF. The synergistic approach yields superior results, especially on reasoning.}
\label{tab:rlhf_results}
\end{table}

\section{Conclusion}
In this paper, we proposed Low-Confidence Gold (LCG), a novel data filtering framework that combines cluster-centric pseudo-labeling with early-stopped classifier training for efficient instruction tuning. Through extensive experiments, we demonstrated the strong performance across multiple benchmarks and base models, validating the effectiveness of our semi-supervised learning paradigm in maintaining both data quality and diversity for instruction tuning.

\section{Limitation}
Our work introduces a semi-supervised training paradigm to curate a subset of data for instruction tuning based on confidence score. However, there still exist several challenges: 1) Even though classifiers are tiny and spend low computational resources to train, it still takes time and effort to initially select data with annotated pseudo-labels. 2) It is likely to be hindered by the original biases and tasks of the dataset, which might still cause inefficiency after selection.

\bibliography{custom}

@misc{alpaca,
  author = {Rohan Taori and Ishaan Gulrajani and Tianyi Zhang and Yann Dubois and Xuechen Li and Carlos Guestrin and Percy Liang and Tatsunori B. Hashimoto },
  title = {Stanford Alpaca: An Instruction-following LLaMA model},
  year = {2023},
  publisher = {GitHub},
  journal = {GitHub repository},
  howpublished = {\url{https://github.com/tatsu-lab/stanford_alpaca}},
}

@misc{mergeit,
      title={MergeIT: From Selection to Merging for Efficient Instruction Tuning}, 
      author={Hongyi Cai and Yuqian Fu and Hongming Fu and Bo Zhao},
      year={2025},
      eprint={2503.00034},
      archivePrefix={arXiv},
      primaryClass={cs.LG},
      url={https://arxiv.org/abs/2503.00034}, 
}

@inproceedings{ge2024clustering,
 title={Clustering and Ranking: Diversity-preserved Instruction Selection through Expert-aligned Quality Estimation},
 author={Ge, Yixuan and Liu, Yang and Hu, Chi and Meng, Weijie and Tao, Shengxuan and Zhao, Xiaopu and Ma, Haoran and Zhang, Liang and Chen, Boxing and Yang, Hongfei and Li, Bei and Xiao, Tong and Zhu, Jingbo},
 booktitle={Proceedings of the 2024 Conference on Empirical Methods in Natural Language Processing},
 pages={464--478},
 year={2024},
 organization={Association for Computational Linguistics}
}

@misc{gsm8k,
      title={Training Verifiers to Solve Math Word Problems},
      author={Karl Cobbe and Vineet Kosaraju and Mohammad Bavarian and Jacob Hilton and Reiichiro Nakano and Christopher Hesse and John Schulman},
      year={2021},
      eprint={2110.14168},
      archivePrefix={arXiv},
      primaryClass={cs.LG}
}

@inproceedings{hu2021lora,
  title={LoRA: Low-Rank Adaptation of Large Language Models},
  author={Hu, Edward J. and Shen, Yelong and Wallis, Phillip and Allen-Zhu, Zeyuan and Li, Yuanzhi and Wang, Shean and Chen, Weizhu},
  booktitle={International Conference on Learning Representations},
  year={2022}
}

@article{jiang2023mistral,
  title={Mistral 7B},
  author={Jiang, Albert Q. and Sablayrolles, Alexandre and Mensch, Arthur and Bamford, Charlie and Chaplot, Devendra Singh and De Las Casas, Diego and Bressand, Florian and Lengyel, Gianna and Lample, Guillaume and Saulnier, Lucile and Lavaud, Leo Raymond and Lachaux, Marie-Anne and Stock, Pierre and Scao, Teven Le and Lavril, Thomas and Wang, Tao and Lacroix, Thibaut and El Sayed, Wissam},
  journal={arXiv preprint arXiv:2310.06825},
  year={2023},
  month={10}
}

@article{sanh2019distilbert,
  title={DistilBERT, a distilled version of BERT: smaller, faster, cheaper and lighter},
  author={Sanh, Victor and Debut, Lysandre and Chaumond, Julien and Wolf, Thomas},
  journal={arXiv preprint arXiv:1910.01108},
  year={2019}
}

@article{scikit-learn,
  title={Scikit-learn: Machine Learning in {P}ython},
  author={Pedregosa, F. and Varoquaux, G. and Gramfort, A. and Michel, V. and Thirion, B. and Grisel, O. and Blondel, M. and Prettenhofer, P. and Weiss, R. and Dubourg, V. and Vanderplas, J. and Passos, A. and Cournapeau, D. and Brucher, M. and Perrot, M. and Duchesnay, E.},
  journal={Journal of Machine Learning Research},
  volume={12},
  pages={2825--2830},
  year={2011}
}

@article{deita,
  title={What makes good data for alignment? A comprehensive study of automatic data selection in instruction tuning},
  author={Liu, Wenxuan and Zeng, Weiwen and He, Kaiyan and Jiang, Yijun and He, Jun},
  journal={arXiv preprint arXiv:2312.15685},
  year={2023},
  month={12}
}

@inproceedings{LIMA,
  title={Long Is More for Alignment: A Simple but Tough-to-Beat Baseline for Instruction Fine-Tuning},
  author={Zhao, Hao and Andriushchenko, Maksym and Croce, Francesco and Flammarion, Nicolas},
  booktitle={International Conference on Machine Learning},
  series={Proceedings of Machine Learning Research},
  year={2024},
  publisher={PMLR}
}

@inproceedings{wang2022minilmv2,
    title={MinILMv2: Multi-Task Pre-training for Multi-Task All-Purpose Text Representations},
    author={Wang, Wenhui and Dong, Li and Cheng, Hao and Wei, Furu and Zhou, Ming},
    booktitle={Findings of the Association for Computational Linguistics: ACL 2022},
    pages={2907--2918},
    year={2022}
}

@inproceedings{
    chen2023alpagasus,
    title={AlpaGasus: Training a Better Alpaca with Fewer Data},
    author={Lichang Chen and Shiyang Li and Jun Yan and Hai Wang and Kalpa Gunaratna and Vikas Yadav and Zheng Tang and Vijay Srinivasan and Tianyi Zhou and Heng Huang and Hongxia Jin},
    booktitle={The Twelfth International Conference on Learning Representations},
    year={2024},
    url={https://openreview.net/forum?id=FdVXgSJhvz}
}

@inproceedings{hellaswag,
    title = "{H}ella{S}wag: Can a Machine Really Finish Your Sentence?",
    author = "Zellers, Rowan  and
      Holtzman, Ari  and
      Bisk, Yonatan  and
      Farhadi, Ali  and
      Choi, Yejin",
    booktitle = "Proceedings of the 57th Annual Meeting of the Association for Computational Linguistics",
    month = jul,
    year = "2019",
    address = "Florence, Italy",
    publisher = "Association for Computational Linguistics",
    url = "https://www.aclweb.org/anthology/P19-1472",
    doi = "10.18653/v1/P19-1472",
    pages = "4791--4800",
}

@misc{arc,
      title={Think you have Solved Question Answering? Try ARC, the AI2 Reasoning Challenge}, 
      author={Peter Clark and Isaac Cowhey and Oren Etzioni and Tushar Khot and Ashish Sabharwal and Carissa Schoenick and Oyvind Tafjord},
      year={2018},
      eprint={1803.05457},
      archivePrefix={arXiv},
      primaryClass={cs.AI}
}

@inproceedings{
mmlu,
title={Measuring Massive Multitask Language Understanding},
author={Dan Hendrycks and Collin Burns and Steven Basart and Andy Zou and Mantas Mazeika and Dawn Song and Jacob Steinhardt},
booktitle={International Conference on Learning Representations},
year={2021},
url={https://openreview.net/forum?id=d7KBjmI3GmQ}
}

@article{dubey2024llama,
  title={The llama 3 herd of models},
  author={Dubey, Abhimanyu and Jauhri, Abhinav and Pandey, Abhinav and Kadian, Abhishek and Al-Dahle, Ahmad and Letman, Aiesha and Mathur, Akhil and Schelten, Alan and Yang, Amy and Fan, Angela and others},
  journal={arXiv preprint arXiv:2407.21783},
  year={2024}
}

@inproceedings{li2024superfiltering,
    title = "Superfiltering: Weak-to-Strong Data Filtering for Fast Instruction-Tuning",
    author = "Li, Ming  and
      Zhang, Yong  and
      He, Shwai  and
      Li, Zhitao  and
      Zhao, Hongyu  and
      Wang, Jianzong  and
      Cheng, Ning  and
      Zhou, Tianyi",
    editor = "Ku, Lun-Wei  and
      Martins, Andre  and
      Srikumar, Vivek",
    booktitle = "Proceedings of the 62nd Annual Meeting of the Association for Computational Linguistics (Volume 1: Long Papers)",
    month = aug,
    year = "2024",
    address = "Bangkok, Thailand",
    publisher = "Association for Computational Linguistics",
    url = "https://aclanthology.org/2024.acl-long.769",
    pages = "14255--14273",
}

@inproceedings{gpt,
 author = {Brown, Tom and Mann, Benjamin and Ryder, Nick and Subbiah et al.},
 booktitle = {Advances in Neural Information Processing Systems},
 pages = {1877--1901},
 title = {Language Models are Few-Shot Learners},
 volume = {33},
 year = {2020}
}

@article{xu2024magpie,
title={Magpie: Alignment Data Synthesis from Scratch by Prompting Aligned {LLMs} with Nothing},
author={Xu, Zhiqing and Jiang, Fanjia and Niu, Lu and Deng, Yiming and Poovendran, Radha and Choi, Yejin and Lin, Bill Yuchen},
journal={arXiv preprint arXiv:2406.08464},
year={2024},
month={6}
}

@article{panickssery2024llm,
  title={LLM evaluators recognize and favor their own generations},
  author={Panickssery, Athul and Bowman, Samuel R. and Feng, Shangmin},
  journal={arXiv preprint arXiv:2404.13076},
  year={2024},
  month={4},
  url={https://arxiv.org/abs/2404.13076}
}

@inproceedings{liu-etal-2022-makes,
    title = "What Makes Good In-Context Examples for {GPT}-3?",
    author = "Liu, Jiachang  and
      Shen, Dinghan  and
      Zhang, Yizhe  and
      Dolan, Bill  and
      Carin, Lawrence  and
      Chen, Weizhu",
    booktitle = "Proceedings of Deep Learning Inside Out (DeeLIO 2022): The 3rd Workshop on Knowledge Extraction and Integration for Deep Learning Architectures",
    year = "2022",
    pages = "100--114",
}

@article{longpre2023flan,
  title={The flan collection: Designing data and methods for effective instruction tuning},
  author={Longpre, Shayne and Hou, Le and Vu, Tu and Webson, Albert and Chung, Hyung Won and Tay, Yi and Zhou, Denny and Le, Quoc V and Zoph, Barret and Wei, Jason and others},
  journal={arXiv preprint arXiv:2301.13688},
  year={2023}
}

\newpage
\appendix

\section{Extended Analysis of Semi-Supervised Model Configurations}
\label{sec:appendix}

\subsection{MultinomialNB Implementation}
\begin{figure*}[t]
    \centering
    \includegraphics[width=0.9\linewidth]{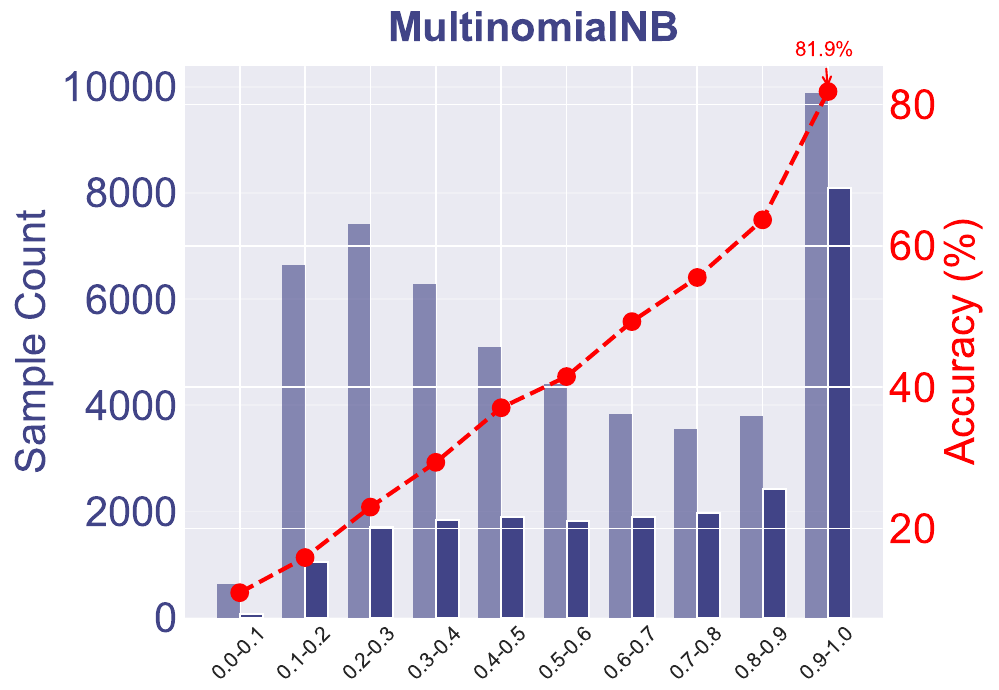}
    \caption{The data distribution of MultinomialNB across different confidence intervals.}
    \label{fig:mul}
\end{figure*}

The confidence distribution patterns of our MultinomialNB baseline, as visualized in Fig. \ref{fig:mul}, reveal fundamentally different characteristics compared to deep learning architectures. The histogram demonstrates remarkable uniformity across confidence intervals (0.0-1.0 with 0.1 increments), showing no significant concentration in specific confidence ranges. This equilibrium phenomenon stems from the model's inherent probabilistic nature and linear decision boundaries, which produce well-calibrated confidence estimates despite its simplicity. 

\subsection{DistilBERT comparative experiment on learning rate}
Our DistilBERT implementation employed a systematic exploration of learning rate hyperparameters {1e-4, 1e-5, 1e-6} within the following experimental framework:
\begin{enumerate}
    \item Architecture: DistilBERT-base-uncased (66M parameters) with custom classification head.
    \item Optimization: Adam optimizer.
    \item Training regime: 3-epoch constraint to prevent overfitting in low-data scenarios.
    \item Data alignment: Identical train/test splits (stratified sampling) as MultinomialNB for direct comparability.
\end{enumerate}

\begin{figure*}[t]
    \centering
    \includegraphics[width=0.9\linewidth]{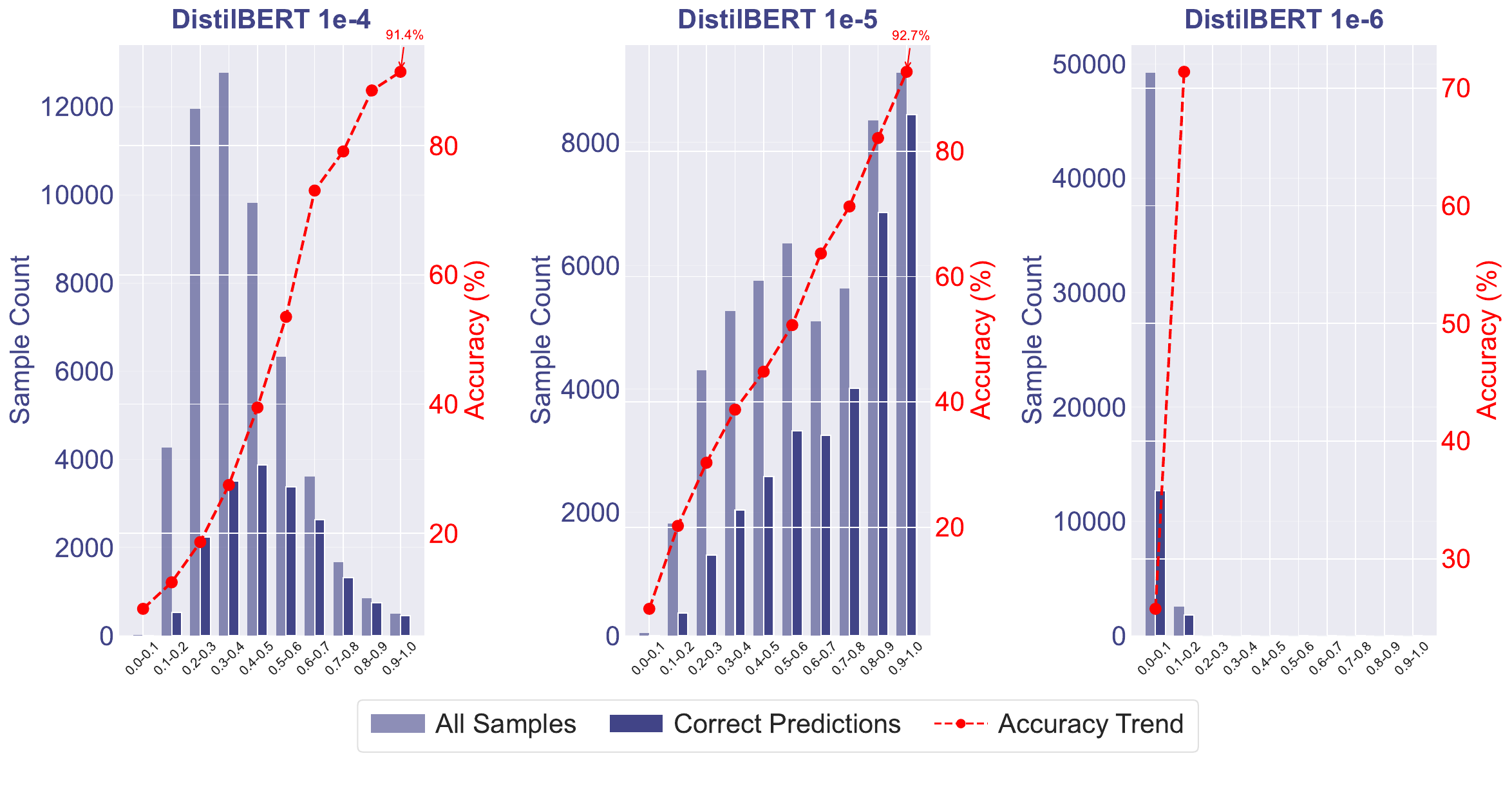}
    \caption{The data distribution of DistilBERT across different confidence intervals under various learning rates.}
    \label{fig:dis}
\end{figure*}

The empirical results (shown in Fig. \ref{fig:dis}) demonstrate non-monotonic performance relationships with learning rate scaling. Peak accuracy (62\%) emerged at 1e-5, while extreme values at both ends (1e-4: 36\%, 1e-6: 28\%) showed substantial performance degradation. This U-shaped accuracy curve suggests the existence of optimal learning rate basins in semi-supervised BERT fine-tuning.

The model exhibited distinct confidence distribution characteristics at the 1e-6 learning rate, with predictions predominantly clustered in the low-confidence range (0-0.2). However, as revealed in Figure 2, comparative analysis across learning rates demonstrated minimal performance variation, showing only marginal improvements that correlated with accuracy increments.


\end{document}